# MetaSegNet: Metadata-collaborative Vision-Language Representation Learning for Semantic Segmentation of Remote Sensing Images

Libo Wang, Sijun Dong, Ying Chen, Xiaoliang Meng, Shenghui Fang and Songlin Fei

*Abstract*—Semantic segmentation of remote sensing images plays a vital role in a wide range of Earth Observation applications, such as land use land cover mapping, environment monitoring, and sustainable development. Driven by rapid developments in artificial intelligence, deep learning (DL) has emerged as the mainstream for semantic segmentation and has achieved many breakthroughs in the field of remote sensing. However, most DL-based methods focus on unimodal visual data while ignoring rich multimodal information involved in the real world. Non-visual data, such as text, can gather extra knowledge from the real world, which can strengthen the interpretability, reliability, and generalization of visual models. Inspired by this, we propose a novel metadata-collaborative segmentation network (MetaSegNet) that applies vision-language representation learning for semantic segmentation of remote sensing images. Unlike the common model structure that only uses unimodal visual data, we extract the key characteristic (e.g. the climate zone) from freely available remote sensing image metadata and transfer it into geographic text prompts via the generic ChatGPT. Then, we construct an image encoder, a text encoder, and a crossmodal attention fusion subnetwork to extract the image and text feature and apply image-text interaction. Benefiting from such a design, the proposed MetaSegNet not only demonstrates superior generalization in zero-shot testing but also achieves competitive accuracy with the state-of-the-art semantic segmentation methods on the large-scale OpenEarthMap dataset (70.4% mIoU) and the Potsdam dataset (93.3% mean F1 score) as well as the LoveDA dataset (52.0% mIoU).

## I. INTRODUCTION

With the advances in sensor technology, a large variety of remote sensing data has been captured increasingly across the globe. The increasing remote sensing of big data not only greatly advances Earth Observation, but also poses huge challenges to interpretation methods [1]. As one of the most effective interpretation methods, semantic segmentation, the task of pixel-level classification, leads to various Earth Observation applications [2], including land use land cover mapping [3], cloud detection [4], [5], forest and wetland inventory [6]–[8], cropland monitoring [9], building extraction [10] and urban sustainable development [11].

The last decade has witnessed the success of deep learning [12] in semantic segmentation. Since the development of the fully convolutional network (FCN) [13], convolutional neural networks (CNN) with automated feature extraction and powerful representation capabilities, have gradually replaced traditional machine learning methods and become the mainstream of semantic segmentation. For several years, many researchers have focused on developing CNN-based semantic segmentation methods for remote sensing images, which greatly promoted the development of this field [14]. Maggiori et al. [15] developed an end-to-end CNN for large-scale remote sensing image segmentation and demonstrated its excellent capacity in land use land cover (LULC) mapping. Diakogiannis et al. [16] combined the advantages of the encoder-decoder structure and residual connections for network design, achieving excellent performance in semantic segmentation. However, the convolution operation with a local receptive view demonstrates the inability to capture long-range dependencies, leading to limitations in fine-grained semantic segmentation [17]. To address this issue, some researchers have started to integrate various attention mechanisms into networks, such as spatial attention and self-attention, and develop attention-fused

This work was supported by the Natural Science Foundation of Jiangsu Province，China (Grant No.SBK20240698), and partly by the National Natural Science Foundation of China (Grant No.41971352), the Natural Science Foundation of the Jiangsu Higher Education Institutions of China (Grant No.23KJB420005) and the Foundation of the Anhui Basic Surveying and Mapping Information Center (Grant No.2023-K-2) *(Corresponding author: Xiaoliang Meng)*

Libo Wang is with School of Remote Sensing and Geomatics Engineering, Nanjing University of Information Science and Technology, and with Technology Innovation Center of Integration Applications in Remote Sensing and Navigation, Ministry of Natural Resources, P.R. China and also with Jiangsu Engineering Center for Collaborative Navigation/Positioning and Smart Applications, Nanjing 210044, China. (e-mail: rosswanglibo@gmail.com).

Sijun Dong, Xiaoliang Meng and Shenghui Fang are with School of Remote Sensing and Information Engineering, Wuhan University, Wuhan 430079, China.

Ying Chen is with the China Telecom Research Institute, Shanghai 200120, China, and also with the School of Remote Sensing and Information Engineering, Wuhan University, Wuhan 430079, China

Songlin Fei is with Forestry and Natural Resources, Purdue University, West Lafayette, IN 47907, USA.

CNNs [18], [19]. Meanwhile, other researchers have applied Transformers [20] for long-range dependencies modeling of visual features [21].

The above CNN- and transformer-based methods have greatly improved visual feature learning for semantic segmentation of remote sensing images. Nevertheless, relying on the unimodal inference pattern, these methods draw limitations in guaranteeing reliable classification results. To this end, several attempts were made to introduce multimodal data as auxiliary information to strengthen model reliability [22]. Common schemes include 1) integrating the digital surface model (DSM) with optical remote sensing images to compensate for height information of geo-objects in 2D images [23], 2) fusing synthetic aperture radar and optical data to mitigate the effects of clouds [24], and 3) Merging 3D point cloud with RGB images to supplement color and texture information [25].

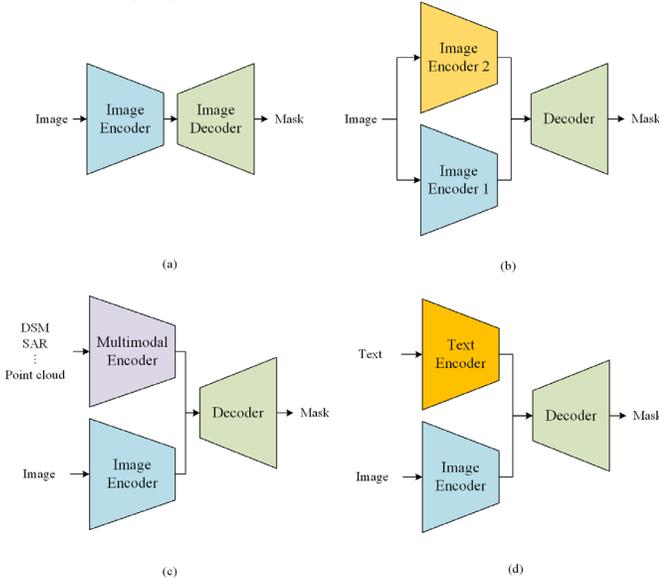

Fig. 1. A review of network architectures in semantic segmentation.

To summarize, the main architectures of remote sensing semantic segmentation networks include (1) the standard encoder-decoder architecture that ensures to output of fine-grained segmentation results, as shown in Fig.1 (a); (2) the dual-path architecture that captures global context and spatial details simultaneously, as shown in Fig.1 (b); (3) the multimodal architecture that introduces multimodal information for complementary, as shown in Fig.1 (c). Moreover, thanks to the great achievements of large language models (LLM), vision-language representation learning or the vision language model (VLM) is becoming a hot topic in the field of computer vision and remote sensing [26]–[32]. In particular, a novel vision-language architecture was applied in semantic segmentation [33], [34], as shown in Fig.1 (d). This architecture usually consists of a text encoder and an image encoder as well as a decoder for image-text feature fusion. Since text can gather extra information or knowledge from the real world [35], the vision-language architecture demonstrates more reliability than vision-only models [36]. As for remote sensing images, text information can describe the meta characteristics of images (e.g. resolution, acquisition time, terrain, and climate zone) or introduce geographic knowledge, thereby promoting an intelligent interpretation of remote sensing images [27].

Inspired by this, we aim to mine the metadata of remote sensing images and introduce knowledge-based geographic text prompts for vision-language representation learning, thereby achieving more precise and reliable segmentation results. Specifically, we propose a **meta**data-collaborative **seg**mentation **net**work, namely MetaSegNet. The MetaSegNet adopts a multimodal vision-language architecture consisting of a text encoder, an image encoder, and a specifically designed crossmodal attention fusion decoder (CAFDecoder). The text encoder applies the universal language transformer (BERT) [37] to encode text features from text prompts. Notably, we employ ChatGPT [38] to generate knowledge-based geographic text prompts. The image encoder applies the general Swin Transformer [39] for image feature extraction. Finally, the CAFDecoder integrates the text and image features and outputs segmentation masks. The main contributions of this paper can be summarized as follows:

(1) We develop a novel metadata-collaborative inference framework with a multimodal vision-language architecture. To the best of our knowledge, we first mine the freely available image metadata and transfer it into knowledge-based geographic text prompts for boosting the reliability of remote sensing segmentation models.

(2) We design a ChatGPT-based pipeline that can generate professional geographic text prompts by querying the characteristics of geo-objects under the specific climate zone extracted from the image metadata.

(3) We design a plug-and-play crossmodal attention fusion decoder that integrates heterogeneous image-text features effectively and strengthens the intra- and inter-modality dependencies to enrich high-level semantic contents.

The remainder of this paper is organized as follows. In Section 2, we present the structure of the proposed MetaSegNet, the inference process, and the developed crossmodal attention fusion decoder. In Section 3, we conduct an ablation study to demonstrate the effectiveness of the CAFDecoder, the impacts of text prompts as well as the zero-shot generalization ability of vision-language architecture and compare the results with a set of advanced segmentation models on three public datasets. In Section 4, we provide a comprehensive discussion. Section 5 is a summary and conclusion.

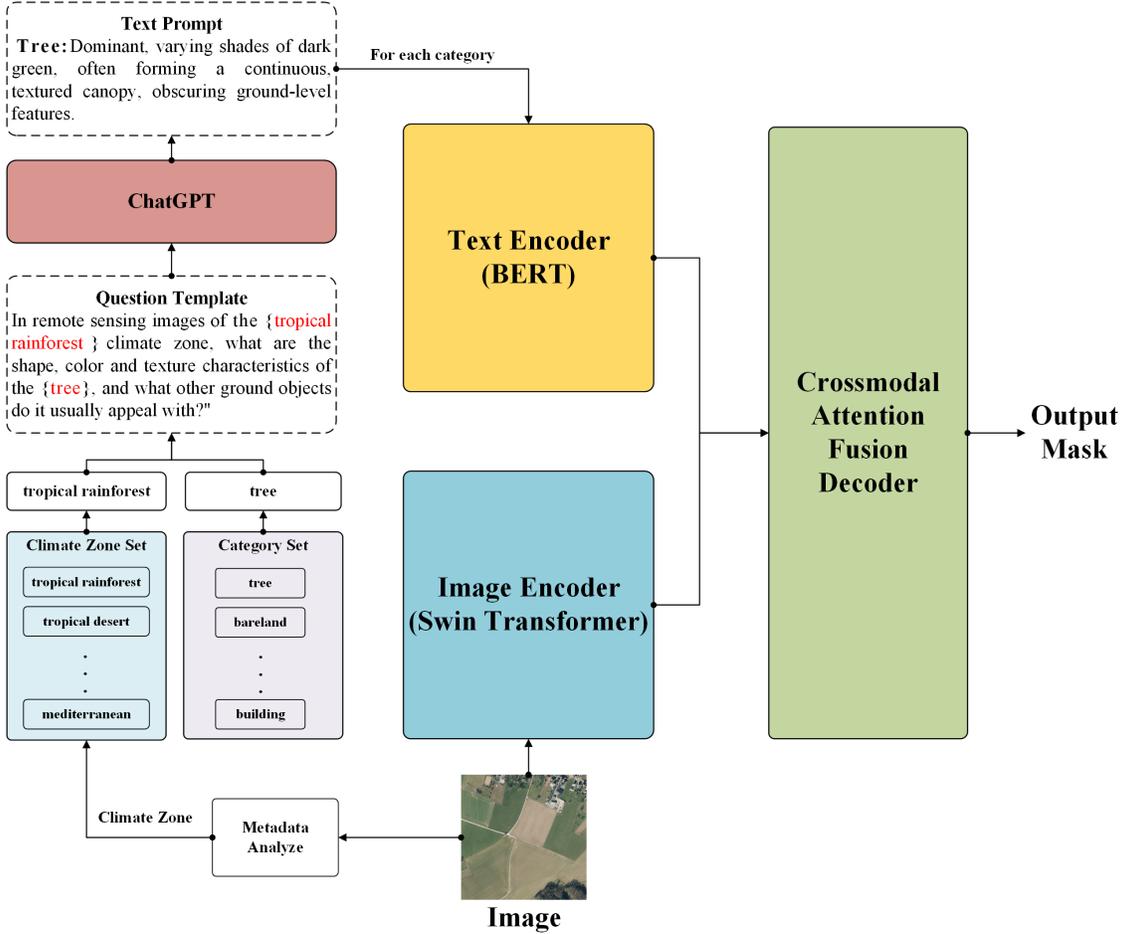

Fig. 2 An overview of our MetaSegNet.

## II. METHODOLOGY

### A. Overview

As illustrated in Fig. 2, the proposed MetaSegNet is constructed using an image encoder, a text encoder, and a crossmodal attention fusion decoder. Besides, a ChatGPT pipeline is also an important part of converting the image metadata to text prompts. We select the base version of Swin Transformer as the image encoder since it allows large-sized input with a low GPU memory requirement. Meanwhile, we choose the general language transformer BERT as the text encoder. Finally, the extracted image and text features are integrated with a crossmodal attention fusion decoder to generate segmentation masks. Detailed descriptions of each component are given in the following sections.

### B. ChatGPT-based text prompt generation

Large language models (LLMs) have demonstrated effectiveness and generalizability simultaneously in a wide range of practical applications, such as question answering and text generation. Following this trend, we apply the well-known ChatGPT to generate geographic text prompts from image metadata. As shown in Fig.1, the main steps for text prompt generation are as follows: 1) We extract the key attributes (e.g. geographical coordinates or regions) from the image metadata.

2) We obtain the climate zone to which the remote sensing image belongs by querying the world map of Koppen-Geiger climate zones [40]. 3) For each category, we generate a question by the designed question template and then ask the ChatGPT to get a text prompt. 4) Finally, we merge each text prompt to generate a complete one and feed it into the text encoder. Benefiting this ChatGPT-based text prompt generation, each remote sensing image is equipped with a detailed description of the shape, color, and texture characteristics of each category as well as potential nearby geo-objects.

### C. Crossmodal attention fusion decoder

The interaction between vision and language frequently occurs in our brains which helps us make sense of this multimodal world. With the success of large-scale multimodal pretraining, vision-language models have become an increasingly central solution for modern AI research [41]. Due to the existing large gap between image and text data, crossmodal image-text feature alignment and fusion are the key factors for vision-language representation learning [42]. To address this issue, we developed the crossmodal attention fusion decoder (CAFDecoder) which consists of three main components, i.e. the image-text matching loss, the image-text alignment module, and the image-text fusion module, as shown in Fig. 3.

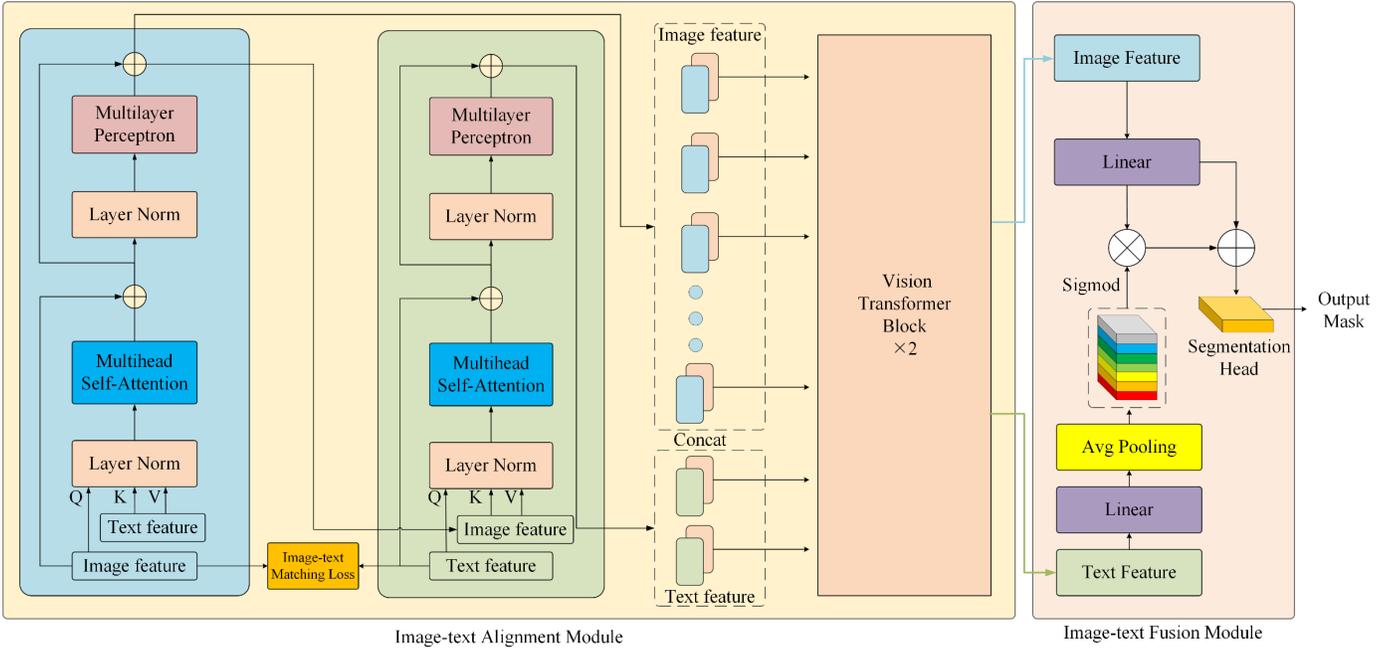

Fig. 3 The architecture of the crossmodal attention fusion decoder.

*1) Image-text alignment module*: We adopt the "Alignment before fusion" strategy to process the heterogeneous image-text feature. We first apply an image-text matching loss to reduce the inter-modal gap between the image feature and the text feature. Such an operation can generate an initial relationship between the two features and promote image-text feature fusion in subsequent processing. Details of the image-text matching loss are as follows:

---

**Algorithms**: Image-text matching loss

---

**Input**: $F_{img}, F_{text}$ #The extracted features of the image encoder and text encoder, respectively.

**Output**: The image-text matching loss $\mathcal{L}_{itm}$.

1. Apply projection and average function on $F_{img}$ and $F_{text}$ to unify the shape $\mathbb{R}^{\in B \times C}$, B denotes the batch size, and C is the channel dimension.
2. Create an image buffer queue and a text buffer queue to save $F_{img}$ and $F_{text}$ of each batch. The queue size is set to 16.
3. Compute the similarity of ($F_{img}$, text queue) and ($F_{text}$, image queue), maximize the image-text pair ($F_{img}$, $F_{text}$) and minimize the other pairs, the same for ($F_{text}$, image queue), then add these two similarities to obtain a contrastive loss.
4. Employ the text encoder BERT to output a positive vector and a negative vector, then use them to obtain a cross-entropy loss.
5. Merge the contrastive loss and cross-entropy loss to output the image-text matching loss $\mathcal{L}_{itm}$.

---

Finally, the total loss $\mathcal{L}$ used for training is a sum of this image-text matching loss $\mathcal{L}_{itm}$, a dice loss $\mathcal{L}_{dice}$ and a cross-entropy loss $\mathcal{L}_{ce}$, whose formulas are as follows:

$$\mathcal{L}_{ce} = -\frac{1}{N}\sum_{n=1}^{N}\sum_{k=1}^{K} y_k^{(n)} \log \hat{y}_k^{(n)} \tag{1}$$

$$\mathcal{L}_{dice} = 1 - \frac{2}{N}\sum_{n=1}^{N}\sum_{k=1}^{K} \frac{\hat{y}_k^{(n)} y_k^{(n)}}{\hat{y}_k^{(n)} + y_k^{(n)}} \tag{2}$$

$$\mathcal{L} = \mathcal{L}_{ce} + \mathcal{L}_{dice} + \mathcal{L}_{itm}(F_{img}, F_{text}) \tag{3}$$

Here, $F_{img} \in \mathbb{R}^{B \times L_1 \times C}$ and $F_{text} \in \mathbb{R}^{B \times L_2 \times C}$ denote the image and the text feature, respectively. B and $C$ are the batch size and the number of channels. $L_1$ and $L_2$ denote the sequence length of the image feature and text feature. In Eq. (1) and (2), $N$ and $K$ represent the number of samples and categories, respectively. $y^{(n)}$ and $\hat{y}^{(n)}$ denote the one-hot vector of the ground truth labels and the corresponding softmax output, $n \in [1, \cdots, N]$. $\hat{y}_k^{(n)}$ is the confidence of sample $n$ belonging to the category $k$.

Then, we perform two consecutive crossmodal attention fusion operations which use the image feature and text feature as the query vector respectively, as shown in **Fig. 3**. Benefiting from such an operation, the inter-modality relations are enhanced between the image and text feature. To further strengthen the inner-modality interaction, we adopt the image-text joint learning scheme which first combines the image and text feature into a position-aware 1D sequence and then uses two standard vision transformers for encoding. The formulas of the two consecutive crossmodal attention fusion operations are as follows:

$$F_{img}^{out} = F_{img}' + \text{MLP}\left(\phi(F_{img}')\right) \tag{4}$$

$$F'_{img} = F_{img} + \text{softmax}\left(\frac{\delta_q(F_{img})\rho_k(F_{text})^T}{\sqrt{d}}\right)\rho_v(F_{text}) \quad (5)$$

$$F^{out}_{text} = F'_{text} + \text{MLP}\left(\emptyset(F'_{text})\right) \quad (6)$$

$$F^{out}_{text} = F_{text} + \text{softmax}\left(\frac{\delta_q(F_{text})\rho_k(F^{out}_{img})^T}{\sqrt{d}}\right)\rho_v(F^{out}_{img}) \quad (7)$$

Where MLP denotes the multilayer perceptron and $\emptyset$ represents a layer normalization operation. $\delta_q$, $\rho_k$ and $\rho_v$ are projection layers that convert the input image feature or text feature to query, key, and value vector, respectively. $d$ is a scale factor.

*2) Image-text fusion module*: The text feature that contains climatic information and characteristics of geo-objects can be a global prior for crossmodal feature fusion. Thus, we first transfer the text feature into a channel-wise global prior by a linear projection and an average pooling operation, as shown in Fig. 3. Then, we deploy a multiplication operation between the processed text feature and the image feature and a residual connection to accomplish the image-text fusion. Finally, the fused feature is fed into the segmentation head to output masks.

## III. EXPERIMENTAL SETTINGS AND DATASETS

### A. Datasets

*1) OpenEarthMap*: As a large-scale high-resolution land cover mapping dataset [43], the OpenEarthMap dataset is composed of 5000 images with eight land cover classes (bareland, rangeland, developed space, road, tree, water, agriculture land, building) at a 0.25–0.5m spatial resolution, covering 97 regions from 44 countries across six continents. Semantic segmentation of OpenEarthMap is challenging due to its wide regional variation, vague categories, and generally complex scenes. In the OpenEarthMap benchmark dataset, the RGB images from each region were randomly divided into training, validation, and test sets, which yielded 3000, 500, and 1500 images, respectively. Notably, labels of the test set are not yet publicly available, so we used the validation set to verify model performance. In our experiments, the input images were uniformly resized to 1024×1024 px patches, and common augmentation strategies like horizontal and vertical flips were used in the training and testing phase.

*2) LoveDA*: The LoveDA dataset [44] contains 5987 fine-resolution optical remote sensing images (GSD 0.3 m) collected from three cities (Nanjing, Changzhou, and Wuhan) in China. The image size is 1024×1024 and the landcover categories include building, road, water, barren, forest, agriculture, and background. To be specific, 2522 images were used for training, 1669 images for validation, and officially provided 1796 images for testing. The dataset involves two scenes (urban and rural), thus, demonstrating considerable challenges. Similarly, the text prompt was adjusted based on the subtropical monsoon climate zone.

*3) Potsdam*: The Potsdam dataset consists of 38 very fine resolution aerial images (Ground sample distance 5 cm) at a size of 6000×6000 pixels and involves 6-class geo-objects

(impervious surface, low vegetation, tree, car, building, and clutter), four multispectral bands (red, green, blue, and near-infrared), as well as the DSM and NDSM. In our experiments, we followed the official train, validation, and test sets partition, and three bands (red, green, and blue) were used for training and testing. The original image tiles were cropped into 1024×1024 px patches as the input, and we applied random flip augmentation to expand training samples. Besides, Potsdam is located in a temperate continental climate zone, thus, the text prompt was adjusted based on this climate zone.

### B. Evaluation Metrics

We use the overall accuracy (OA), mean intersection over union (mIoU), F1 score, precision, and recall to evaluate the performance of models, which can be defined as follows:

$$OA = \frac{\sum_{k=1}^{N} TP_k}{\sum_{k=1}^{N} TP_k + FP_k + TN_k + FN_k}, \quad (8)$$

$$mIoU = \frac{1}{N}\sum_{k=1}^{N}\frac{TP_k}{TP_k + FP_k + FN_k}, \quad (9)$$

$$precision = \frac{1}{N}\sum_{k=1}^{N}\frac{TP_k}{TP_k + FP_k}, \quad (10)$$

$$recall = \frac{1}{N}\sum_{k=1}^{N}\frac{TP_k}{TP_k + FN_k}, \quad (11)$$

$$F1 = 2 \times \frac{precision \times recall}{precision + recall}, \quad (12)$$

where $TP_k$, $FP_k$, $TN_k$, and $FN_k$ indicate the true positive, false positive, true negative, and false negatives, respectively, for the specific object indexed as class $k$. OA is computed for all categories including the background pixels.

### C. Experimental Setting

All models in the experiments were implemented by the PyTorch framework 2.0 on a single NVIDIA GTX 4090 GPU. We deployed the AdamW optimizer to train all models and the early stopping strategy was used for avoiding overfitting. The init learning rate and batch size were set to 3e-4 and 2, respectively. The weight decay was set to 2.5e-4. The max training epoch was 45 and the cosine strategy was applied to adjust the learning rate. In particular, the text encoder BERT was frozen in the training phase. The input text prompt was padded to a uniform length of 250. Moreover, as for a 1024×1024 px input patch, our MetaSegNet has a competitive FPS of 23.7. The total number of parameters of our MetaSegNet is 399.7 M, of which only a quarter are trainable parameters.

### D. Benchmark Methods

**Baseline**: The baseline consists of the Swin Transformer (Swin-Base) [39] and a segmentation head with a series of upsampling operations, which is a pure vision model constructed for ablation experiments with the proposed MetaSegNet.

**U-Net**: The U-Net [45] is a classic fully convolutional neural network that first introduces the encoder-decoder structure for promoting semantic segmentation.

**DANet**: The DANet [46] is an attention-based convolutional neural network with spatial attention and channel attention,

which explores the potential of self-attention mechanisms early for semantic segmentation.

**SegFormer**: The SegFormer [47] is a fully transformer-based network following the encoder-decoder structure, which improves the transformer architecture for dense prediction.

**CLIPSeg**: The CLIPSeg [33] is a multimodal segmentation network following the vision language structure, which introduces text or image prompts for enhancing semantic segmentation.

**Advanced semantic segmentation methods**: We selected a set of remote sensing and natural image segmentation methods for comprehensive comparisons, including PSPNet [48], DeeplabV3+ [49], BoTNet [50], Segmenter [51], SwinUperNet [39], V-FuseNet [23], UFMG_4 [52], ResUNet-a [16], DDCM-Net [53], LANet [54], FarSeg [55], FactSeg [56], EaNet [57], BANet [58], ABCNet [59], UNetFormer [17], MANet [19], DC-Swin [60], SAPNet [61], EMRT [62], FTransUNet[63], and RS³Mamba[64].

## IV. EXPERIMENTAL RESULTS AND ANALYSIS

### A. Quantitative Comparisons with State-of-the-art Semantic Segmentation Methods

*1) OpenEarthMap*: OpenEarthMap is a large-scale land cover mapping dataset, where the images are captured by fine-resolution satellite or aerial sensors, covering different cities around the world. Therefore, it is very challenging to achieve a high accuracy on this dataset. We trained several advanced semantic segmentation models and provided detailed comparisons on the OpenEarthMap validation set. As illustrated in Table I, the MetaSegNet yields the best mIoU (70.4%) while demonstrating advantages in most of the specific categories. Specifically, our method not only exceeds the CNN-based models (e.g. UNet, DANet, and MANet) by at least 6.4% and the advanced Transformer-based network UNetFormer by 2.4% but also outperforms the recent Mamba-based method RS³Mamba by 5.9%. Furthermore, the visualization results (Fig. 4) also demonstrate the superiority of our MetaSegNet compared to the recent CNNs and vision transformers.

TABLE I

QUANTITATIVE COMPARISONS WITH STATE-OF-THE-ART METHODS ON THE OPENEARTHMAP VALIDATION SET. THE BOLD DENOTES THE BEST VALUES. THE UNDERLINE DENOTES THE SECOND-BEST VALUES.

| Method | Background | Bareland | Rangeland | Developed | Road | Tree | Water | Agriculture | Building | mIoU |
|---|---|---|---|---|---|---|---|---|---|---|
| UNet [45] | 82.0 | 23.6 | 52.3 | 49.8 | 57.3 | 67.2 | 67.1 | 71.7 | 73.0 | 60.4 |
| DANet [46] | 81.1 | 30.9 | 49.8 | 47.5 | 57.4 | 63.3 | 67.6 | 73.2 | 70.2 | 60.1 |
| BoTNet [50] | 81.9 | 30.3 | 50.9 | 50.2 | 57.9 | 67.8 | 65.5 | 75.8 | 72.9 | 61.5 |
| MANet [19] | 91.0 | 41.2 | 50.8 | 50.8 | 50.8 | 60.0 | 69.3 | 70.1 | 75.4 | 64.0 |
| SegFormer [65] | 97.2 | 41.0 | 56.4 | 53.2 | 58.7 | 70.9 | 77.7 | 76.7 | 76.2 | 66.0 |
| DC-Swin [60] | 96.4 | 42.5 | 55.6 | 53.2 | 58.1 | 69.9 | 77.8 | 76.0 | 75.7 | 67.2 |
| UNetFormer [17] | 97.2 | 42.8 | 56.2 | 53.5 | 60.9 | 70.2 | 77.4 | 76.6 | 76.9 | 68.0 |
| CLIPSeg [33] | 95.5 | 34.5 | 46.1 | 40.8 | 45.2 | 60.0 | 69.1 | 73.7 | 62.2 | 58.6 |
| RS³Mamba [64] | 96.3 | 39.9 | 51.0 | 48.7 | 56.9 | 66.8 | 74.4 | 75.0 | 71.4 | 64.5 |
| MetaSegNet (Ours) | **97.5** | **44.3** | **59.3** | **57.5** | **63.7** | **72.2** | **80.7** | **79.6** | **79.0** | **70.4** |

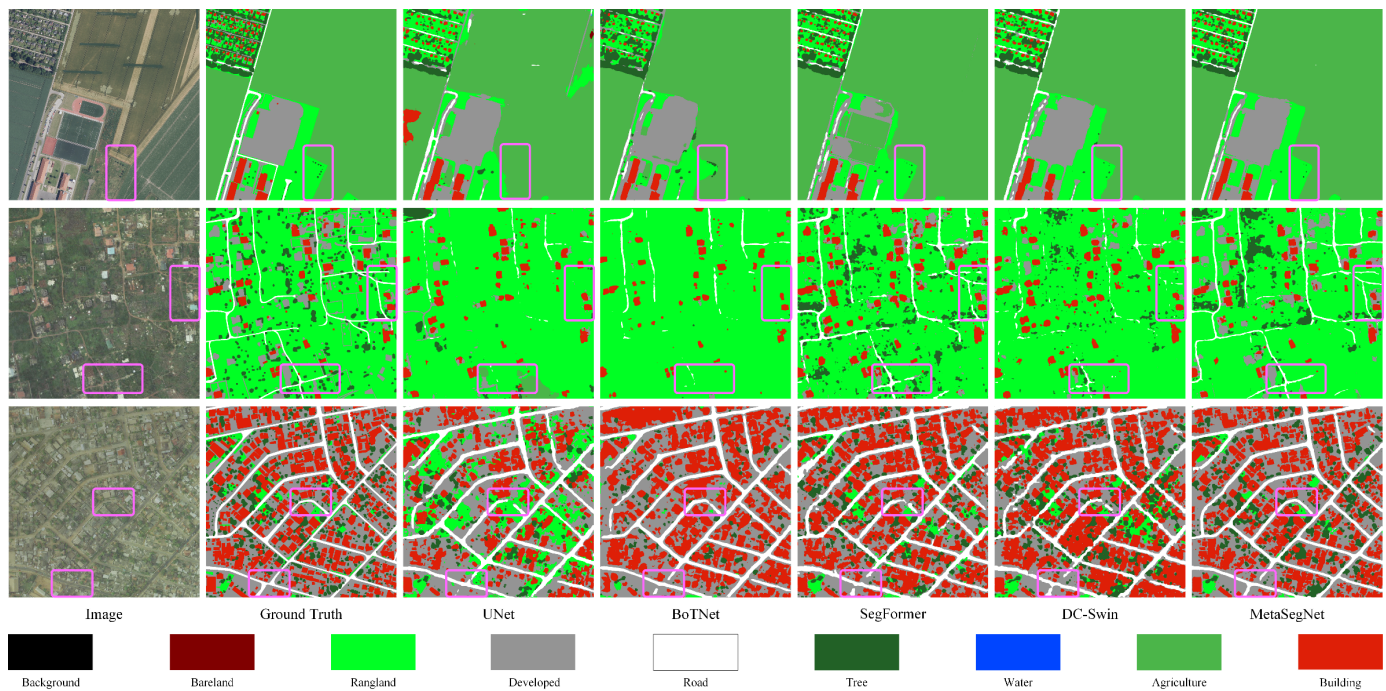

Fig. 4 Visual comparisons on the OpenEarthMap dataset.

*2) Potsdam*: The ISPRS Potsdam is a widely-used dataset for verifying remote sensing image segmentation methods.

Currently, many specially designed networks obtained outstanding performance on this dataset. In this section, we demonstrate that our MetaSegNet can achieve competitive scores in comparison with these state-of-the-art networks. As shown in Table II, our MetaSegNet yields a 93.3% mean F1 score, a 92.0% overall accuracy, and an 87.6% mIoU on the Potsdam test set. The results of the MetaSegNet not only surpass the CNN-based segmentation model EaNet by 2.6% in mean F1 score but also outperform the recent Transformer-based network FTransUNet by 1.4% in mIoU. We also provided visualization comparisons with UNetFormer, as exhibited in Fig. 5, the overall segmentation performance is also excellent.

TABLE II
QUANTITATIVE COMPARISONS WITH STATE-OF-THE-ART METHODS ON THE POTSDAM TEST SET. THE BOLD DENOTES THE BEST VALUES. THE UNDERLINE DENOTES THE SECOND-BEST VALUES.

| Method | Imp. surf. | Building | Low. veg. | Tree | Car | Mean F1 | OA | mIoU |
|---|---|---|---|---|---|---|---|---|
| V-FuseNet [23] | 92.7 | 96.3 | 87.3 | 88.5 | 95.4 | 92.0 | 90.6 | - |
| UFMG_4 [52] | 90.8 | 95.6 | 84.4 | 84.3 | 92.4 | 89.5 | 87.9 | - |
| ResUNet-a [16] | 93.5 | 97.2 | 88.2 | 89.2 | 96.4 | 92.9 | 91.5 | - |
| DDCM-Net [53] | 92.9 | 96.9 | 87.7 | 89.4 | 94.9 | 92.3 | 90.8 | - |
| LANet [54] | 93.1 | 97.2 | 87.3 | 88.0 | 94.2 | 92.0 | 90.8 | - |
| EaNet [57] | 92.0 | 95.7 | 84.3 | 85.7 | 95.1 | 90.6 | 88.7 | 83.4 |
| BANet [58] | 93.3 | 96.7 | 87.4 | 89.1 | 96.0 | 92.5 | 91.0 | 86.3 |
| ABCNet [59] | 93.5 | 96.9 | 87.9 | 89.1 | 95.8 | 92.7 | 91.3 | 86.5 |
| UNetFormer [17] | 93.6 | 97.2 | 87.7 | 88.9 | 96.5 | 92.8 | 91.3 | 86.8 |
| SAPNet [61] | 94.4 | 97.7 | 87.8 | 89.6 | 96.0 | 93.1 | 91.8 | - |
| FTransUNet [63] | 93.1 | 97.7 | 88.4 | 88.2 | 96.3 | 92.4 | - | 86.2 |
| MetaSegNet (ours) | 94.5 | 97.2 | 88.3 | 89.8 | 96.6 | 93.3 | 92.0 | 87.6 |

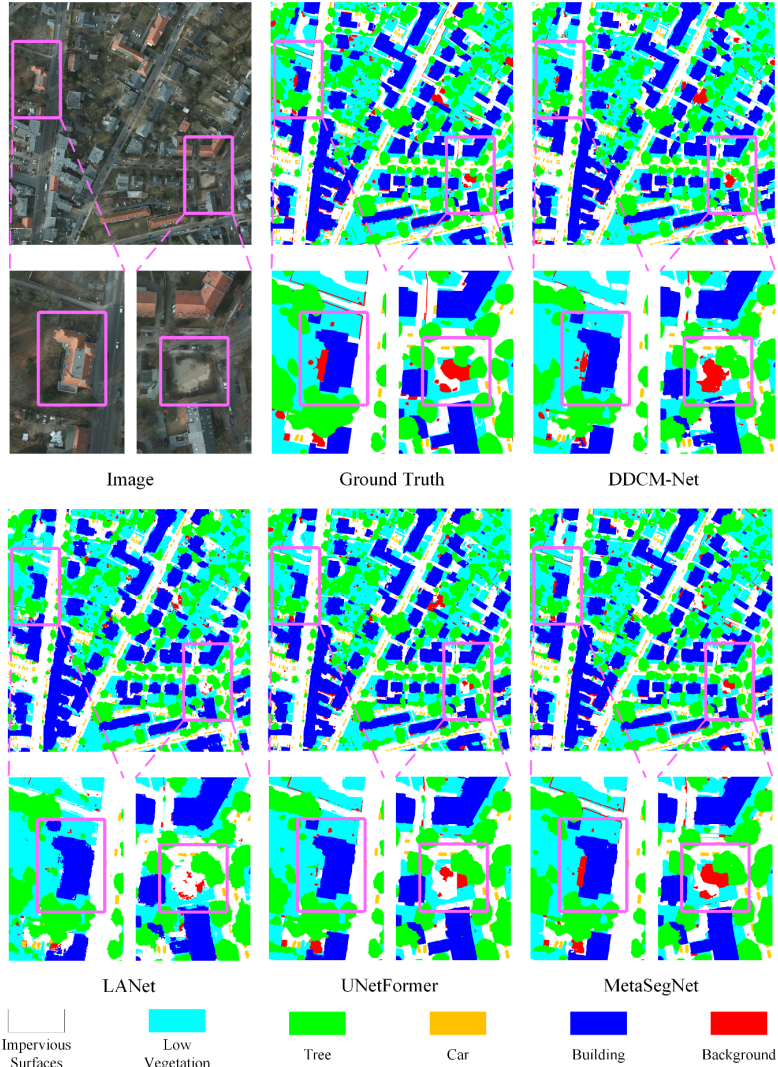

Image     Ground Truth     DDCM-Net

LANet     UNetFormer     MetaSegNet

Impervious Surfaces    Low Vegetation    Tree    Car    Building    Background

Fig. 5 Visual comparisons on the Potsdam dataset.

*3) LoveDA*: We conducted experiments on the LoveDA dataset to further test the proposed MetaSegNet. As listed in Table III, our MetaSegNet obtains the highest mIoU (52.0%). These results indicate the advantages of the vision-language structure and informative text prompts in coping with complex remotely sensed scenes compared to other CNNs and Transformers. Besides, the visualization results on the LoveDA validation set also demonstrate the advantages of our method on specific categories, especially on the roads and buildings, as shown in Fig.6.

TABLE III

QUANTITATIVE COMPARISONS WITH STATE-OF-THE-ART METHODS ON THE LOVEDA TEST SET. THE BOLD DENOTES THE BEST VALUES. THE UNDERLINE DENOTES THE SECOND-BEST VALUES.

| Method | Background | Building | Road | Water. | Barren | Forest | Agriculture | mIoU |
|---|---|---|---|---|---|---|---|---|
| PSPNet [48] | 44.4 | 52.1 | 53.5 | 76.5 | 9.7 | 44.1 | 57.9 | 48.3 |
| DeepLabV3+ [49] | 43.0 | 50.9 | 52.0 | 74.4 | 10.4 | 44.2 | 58.5 | 47.6 |
| FarSeg [55] | 43.1 | 51.5 | 53.9 | 76.6 | 9.8 | 43.3 | 58.9 | 48.2 |
| FactSeg [56] | 42.6 | 53.6 | 52.8 | 76.9 | 16.2 | 42.9 | 57.5 | 48.9 |
| Segmenter [51] | 38.0 | 50.7 | 48.7 | 77.4 | 13.3 | 43.5 | 58.2 | 47.1 |
| SwinUperNet [39] | 43.3 | 54.3 | 54.3 | 78.7 | 14.9 | 45.3 | 59.6 | 50.0 |
| BANet [58] | 43.7 | 51.5 | 51.1 | 76.9 | 16.6 | 44.9 | 62.5 | 49.6 |
| DC-Swin [60] | 41.3 | 54.5 | 56.2 | 78.1 | 14.5 | 47.2 | 62.4 | 50.6 |
| EMRT [62] | 50.5 | 60.4 | 52.7 | 66.7 | 32.8 | 40.2 | 52.5 | 50.8 |
| RS³Mamba [64] | 39.7 | 58.7 | 57.9 | 61.0 | 37.2 | 39.6 | 33.9 | 50.9 |
| MetaSegNet (ours) | 43.5 | 58.7 | 57.8 | 80.8 | 17.2 | 46.6 | 59.1 | 52.0 |

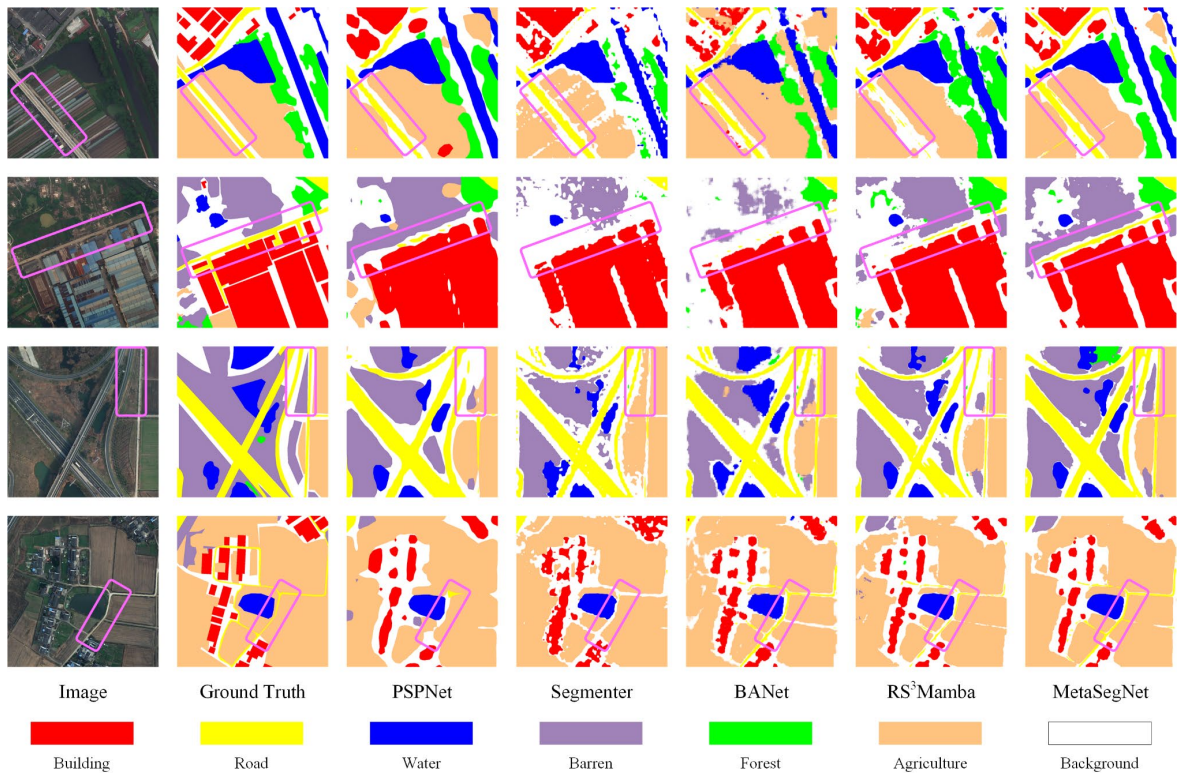

Fig. 6 Visual comparisons on the LoveDA validation set.

representation learning.

## B. Ablation Study

*1) Components of the MetaSegNet*: To evaluate the performance of the image-text alignment module (indicated as Alignment) and image-text fusion module (indicated as Fusion), we conducted an ablation experiment on the OpenEarthMap dataset. The results are shown in Table IV. With the deployment of the image-text alignment module, we obtained an increase of 2.5% mIoU, which demonstrates the effectiveness and necessity of this module for image-text feature fusion. Moreover, the employment of the image-text fusion module can provide an extra improvement of 0.8% mIoU, which indicates its effectiveness in vision-language

TABLE IV

THE ABLATION EXPERIMENTAL RESULTS OF EACH COMPONENT ON THE OPENEARTHMAP DATASET.

| Method | mIoU |
|---|---|
| Baseline | 67.1 |
| Baseline+Alignment | 69.6 |
| Baseline+Alignment+Fusion | 70.4 |

*2) Impacts of text prompts*: To demonstrate the influence of text prompts and the importance of mining image metadata, we replaced the ChatGPT-based text prompts with simple text prompts for ablation studies. As illustrated in Table V, the utilization of a simple text prompt in the MetaSegNet results in a mIoU drop of over 0.5%, which reveals the importance of professional and detailed text prompts. Notably, there is a significant accuracy drop (3.3%) without using any text prompts, which can demonstrate the advantage of vision-language models compared to vision-only models. Moreover, climate zones usually have a more significant impact on vegetation than human-made geo-objects. Thus, we select the tree, agriculture, and building to further reveal the influence of climate-aware text prompts. The results show that the accuracy reduction of buildings (0.3%) is less than that of trees (0.6%) and agricultural land (1.3%) when applying a non-climate text prompt, which can illustrate the importance and advantage of knowledge geographic text prompts and indicate the success of image-text fusion indirectly.

TABLE V
IMPACTS OF TEXT PROMPTS.

| Text Prompt | mIoU | IoU-Building | IoU-Tree | IoU-Agriculture |
|---|---|---|---|---|
| No text prompt | 67.1 | 73.3 | 67.6 | 77.5 |
| "It is a remote sensing image." | 69.9 | 78.7 | 71.6 | 78.3 |
| Our ChatGPT-based text prompt | 70.4 | 79.0 | 72.2 | 79.6 |

*3) Zero-shot testing*: To evaluate the generalization of the vision-language architecture, we selected the Potsdam dataset to compare the zero-shot segmentation ability of the baseline and MetaSegNet. We first pre-trained the MetaSegNet and baseline on the large-scale OpenEarthMap then directly applied it on the Potsdam dataset. Since the categories of these two datasets are different, we only computed the IoU of the overlapping categories, i.e. the human-made building and nature-made tree. As shown in Table VI, our MetaSegNet yields a large increase in IoU-Building (2.9%) and IoU-Tree (4.1%) in comparison with the Baseline, which can illustrate the superior generalization ability of vision-language models than pure vision models.

TABLE VI
ZERO-SHOT TESTING ON THE POTSDAM DATASET.

| Method | IoU-Building | IoU-Tree |
|---|---|---|
| Baseline | 70.8 | 30.8 |
| MetaSegNet | 73.7 | 34.9 |

## V. DISCUSSION

*1) Advantages of vision-language representation learning*: With the success of Vision Transformer and large language models (LLMs), vision-language representation learning is becoming a bright research topic. Inspired by this trend, we develop a MetaSegNet with a vision-language structure, which aims to use knowledge-based geographic text prompts to promote the reliability of remote sensing segmentation models. Our experiment results have demonstrated this point. In particular, we revealed that the climate-related text prompts can improve the accuracy of vegetation classifications (e.g. Trees) more significantly in comparison with human-made buildings. Such a phenomenon becomes more obvious in zero-shot testing, which highlights the importance of text-image association in vision-language models.

*2) Potential of large-scale unsupervised pretraining*: Notably, the text prompts used in MetaSegNet can be fully extracted from freely available remote sensing image metadata. Thus, how to use these free metadata in large-scale unsupervised pretraining is also an interesting research topic. In the future, we will further explore this topic deeply, especially its potential in the training of large remote sensing models, and explore more core attributes of metadata (e.g. time and spatial resolution) and their impacts on specific geo-objects.

*3) Knowledge-based geographic text prompt*: Text is a good carrier of geographical knowledge. It is very promising to obtain knowledge-driven deep models by introducing more geographic knowledge and performing text prompt tuning.

## VI. CONCLUSION

In this paper, we proposed a novel vision-language network (MetaSegNet) for semantic segmentation of remote sensing images. We designed a ChatGPT-based pipeline for knowledge-based geographic text prompts generation and a crossmodal attention fusion subnetwork for image-text feature interaction and fusion. Ablation studies indicate that geographic text prompts can boost the model's reliability, especially on text-related geo-objects. Besides, comprehensive experiments on the OpenEarthMap, Potsdam, and LoveDA datasets further demonstrated the advantages of the proposed method and the effectiveness of vision-language representation learning.